\documentclass{article}

     \usepackage[final,nonatbib]{tackling_climate_workshop_style}

\usepackage[utf8]{inputenc} 
\usepackage[T1]{fontenc}    
\usepackage{hyperref}       
\usepackage{url}            
\usepackage{booktabs}       
\usepackage{amsfonts}       
\usepackage{nicefrac}       
\usepackage{microtype}      
\usepackage{xcolor}         

\usepackage{microtype}
\usepackage{graphicx}
\usepackage{wrapfig}
\usepackage{subfigure}
\usepackage{booktabs} 

\usepackage{hyperref}

\usepackage{amssymb} 
\usepackage{amsmath}
\usepackage{amsthm}
\usepackage{bbm}
\usepackage{dsfont}
\usepackage{graphicx}
\usepackage{color}
\usepackage{subfigure}
\usepackage{float}
\usepackage{xfrac}
\graphicspath{{../figs/},{../GenExamples/}}
\definecolor{Red}{rgb}{1.0,0,0} 

\definecolor{Red}{rgb}{1.0,0,0} 

\newcommand{\comm}[1]{\textcolor{Red}{[#1]}} 

\DeclareMathOperator*{\argmin}{argmin}

\newcommand{\pmm}{$PM_{2.5}\,$}
\newcommand{\nitr}{$NO_{3}\,$}
\newcommand{\amm}{$NH_{4}\,$}

\usepackage{xcolor}
\usepackage{xspace}
\makeatletter
\DeclareRobustCommand\onedot{\futurelet\@let@token\@onedot}
\def\@onedot{\ifx\@let@token.\else.\null\fi\xspace}

\makeatother

\usepackage{xargs}                      
\usepackage[colorinlistoftodos,prependcaption,textsize=tiny]{todonotes}
\newcommandx{\unsure}[2][1=]{\todo[linecolor=red,backgroundcolor=red!25,bordercolor=red,#1]{#2}}
\newcommandx{\change}[2][1=]{\todo[linecolor=blue,backgroundcolor=blue!25,bordercolor=blue,#1]{#2}}
\newcommandx{\info}[2][1=]{\todo[linecolor=OliveGreen,backgroundcolor=OliveGreen!25,bordercolor=OliveGreen,#1]{#2}}
\newcommandx{\improvement}[2][1=]{\todo[linecolor=Plum,backgroundcolor=Plum!25,bordercolor=Plum,#1]{#2}}
\newcommandx{\thiswillnotshow}[2][1=]{\todo[disable,#1]{#2}}
\newcommand\todocomment[1]{\textcolor{red}{#1}}

\title{A hybrid convolutional neural network/active contour approach
to segmenting dead trees in aerial imagery}

\author{%
Jacquelyn A.~Shelton,\quad Przemyslaw Polewski,\quad Wei Yao\\
The Hong Kong Polytechnic University\\
Dept. of Land Surveying and Geo-Informatics\\
\texttt{\{jacquelyn.ann.shelton, przemyslaw.polewski\}@gmail.com, wei.hn.yao@polyu.edu.hk}
\And
Marco Heurich\\ 
Bavarian Forest National Park\\
Dept. for Visitor Management and National Park Monitoring\\
and the Dept. of Wildlife Ecology and Management\\
Albert-Ludwigs-Universität Freiburg, Germany\\
\texttt{marco.heurich@npv-bw.bayern.de}\\
}

\begin{document}

\maketitle

\begin{abstract}
The stability and ability of an ecosystem to withstand climate change is directly linked to its biodiversity.
Dead trees are a key indicator of overall forest health, housing one-third of forest ecosystem biodiversity, and constitute $8\%$ of the global carbon stocks.
They are decomposed by several natural factors, e.g. climate, insects and fungi. 
Accurate detection and modeling of dead wood mass is paramount to understanding forest ecology, the carbon cycle and decomposers. 
We present a novel method to construct precise shape contours of dead trees from aerial photographs 
by combining established convolutional neural networks with a novel active contour model in an energy minimization framework. 
Our approach yields superior performance accuracy over state-of-the-art in terms of precision, recall, and intersection over union of detected dead trees. 
This improved performance is essential to meet emerging challenges caused by climate change (and other man-made perturbations to the systems), particularly to monitor  and estimate carbon stock decay rates, monitor forest health and biodiversity, and the overall effects of dead wood on and from climate change.
\end{abstract}

\section{Introduction}
\label{sec:intro}
\vspace{-.3cm}
With the increasing global interest in understanding and mitigating climate change, 
researchers find themselves presented with new problems.
One such problem is understanding the role and behavior of dead trees in these processes, as they are a key indicator of forest health. 
Forests are a core component in the global carbon cycle and are the most efficient ecosystem on the planet for scrubbing CO2 and returning oxygen to the atmosphere, sequestering as much CO2 as all of the oceans.
Carbon stocks and fluxes in dead wood -- fallen and standing dead trees, branches, and other woody tissues -- are a critical component of forest carbon dynamics \cite{nature_carbon_fractions2021}, constituting $8\%$ of the global forest carbon stocks~\cite{science_carbonStocks}. 
Furthermore, dead wood houses one-third of all forest biodiversity, which is of crucial importance as an ecosystem's bioversity is directly linked to its stability and the ability to withstand climate change.
%
Dead trees are decomposed by several natural factors (including climate, fungi and insects), 
however, the influence of these decomposers as well as the impact of environmental change upon them remains poorly understood.
While initial studies of both insects~\cite{nature_main2021} and
fungi~\cite{Bradford2014,Lustenhouwer11551} have been performed, further studies
are still needed to gain a more holistic understanding.
In particular, there is an increasing need for both larger scale and longitudinal
studies of the impact of dead trees on the ecology of forests, and their interaction with the carbon cycle and decomposers (see e.g. \cite{nature_carbon_fractions2021} and citations within).
These efforts are hindered by a lack of data and tools for processing the data, particularly from \textit{aerial photography}, which offers a good trade-off
between high spatial resolution and cost efficiency, making it ideal for localized studies. In order to address this need
we propose the use of Machine Learning (ML) algorithms to identify the location and shape of dead trees.
Namely, using Computer Vision (CV) ML techniques applied to aerial photos of a forest at multiple time steps a temporal change in tree crowns can be made, providing estimates in decay rates.

The motivation of this work is to develop a method to accurately model and estimate dead wood mass. 
There are however other applications ranging e.g. from tracking the health of a forest by identifying dead and dying trees from invasive insects and disease, to tracking desertification and reforestation after harvesting or wildfires, to the development of algal blooms in the ocean. 
The precise fallen tree maps could be further used as a basis for plant and animal habitat modeling, studies on carbon sequestration as well as soil quality in forest ecosystems.

The method we propose is a hybrid of two convolutional neural networks with a novel active contour model for precise object contour segmentation. 
We use infrared aerial imagery to identify dead vegetation, a widely used technique due to the difference in reflectance caused by differences in chlorophyll in the near-infrared spectral band.
Due to recent improvements in this technology, in specific the increase in resolution, the current existent, non-Machine Learning, methods are unable to provide the highest satisfactory performance.
These discrepancies are then only exaggerated when the amount of available data is drastically increased by the use of unmanned aerial vehicles (UAVs) to collect data more often for the same forest.
The details of the method are as follows: We use leading convolutional network
approaches \textit{U-Net} for instance segmentation to compute class
probability masks of the dead trees and a \textit{Mask R-CNN} (Mask Regional-CNN) to segment the image into components, in particular separating trees from each other.
The Mask R-CNN can successfully identify the number and precise position of the trees.
To further improve the contours of the dead trees we then apply a contour refinement step based on a generalized classical computer vision technique by using simultaneously evolving contours based upon energy functions.

The goal of the present work is to design a cutting edge Machine Learning algorithm for identifying dead trees in a forest, and then determining the shape and location of the dead tree's crown. 
With this information, crucial aspects of carbon decay can be more accurately estimated and predicted. 
Experimental results yield superior performance over conventional instance segmentation methods, reducing the cost of large scale studies allowing for improved understanding of forest  health, and how that is impacted through time by factors such as insects, natural disturbances, and especially climate change.
%
The paper is organized as follows: 
Sec.~\ref{sec:methods} introduces the proposed hybrid method, 
Sec.~\ref{sec:exps} presents experimental results, and finally Sec.~\ref{sec:disc} provides a summary of the work and outlook.

\vspace{-.2cm}
%
%
\section{A hybrid approach to contour modeling of dead trees in aerial images}
\label{sec:methods}
\begin{figure*}[ht!]
\begin{center}
\includegraphics[width=.9\columnwidth]{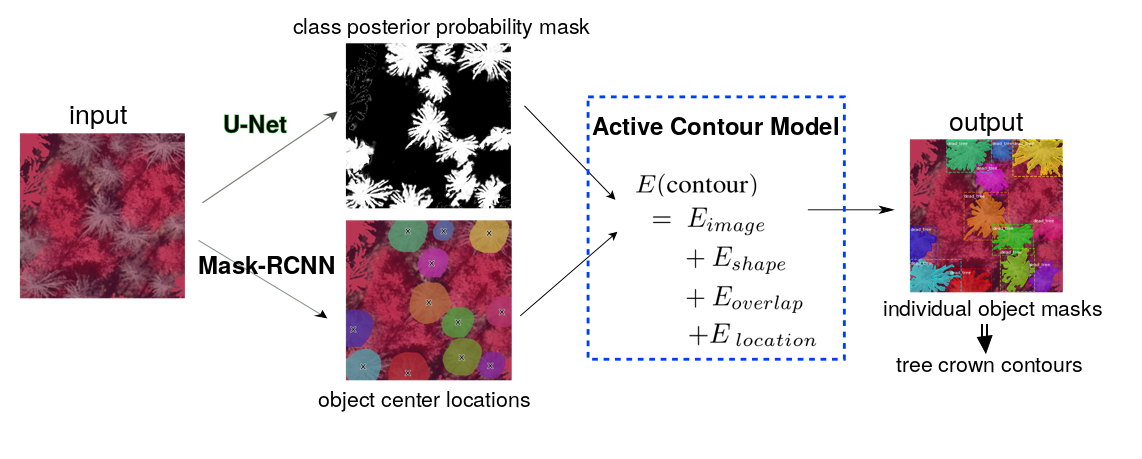}
\vspace{-.6cm}
\caption{Illustration of our strategy for high resolution dead tree contour modeling.}
\label{fig:pipeline}
\end{center}
\end{figure*}

\vspace{-.3cm}
We first describe the \textit{convolutional neural networks} implemented for instance segmentation and object localization.
Then we introduce our \textit{main technical contribution} which harnesses the advantages of these networks to construct our method.
Specifically, the \textit{U-Net} gives us the probabilities of which pixels belong to which dead trees (classes) and the \textit{Mask R-CNN} provides solid estimates of the locations of each tree (centroids).
We combine these results in a \textit{novel energy minimization framework} for high resolution contour modeling.
Figure \ref{fig:pipeline} provides a simplified overview of the entire process.
%

\vspace{-.15cm}
\textbf{U-Net}. 
The U-Net~\cite{unet_paper} is a fully convolutional neural network architecture
which constitutes a milestone in the task of image~\emph{dense semantic
segmentation}.
This network is particularly well suited for our problem because it
preserves the object contours well, an imperative aspect for retaining fine
details of the tree crowns.

\vspace{-.15cm}
\textbf{Mask R-CNN.} Mask R-CNN~\cite{maskRcnnOrig}, or Region Based
Convolutional Neural Networks, is a state of the art neural network architecture
for the task of generic image~\emph{instance segmentation}, i.e.~obtaining
separate pixel masks for all object instances present within the input image.
%
%
Mask R-CNN has two stages: 
(i) a region proposal network, which selects promising image regions that are likely
to contain object instances, and 
(ii) a fine-grained detection component which examines the candidate regions and predicts the object class
label, bounding box, and the instance's pixel mask.


\vspace{-.15cm}
\textbf{Active contour segmentation with energy minimization.}
We formalize the above setting with our contour model as follows. Let
$p^{init}_{i}, i \in \{1,M\}$ be the $M$ object centroids identified by the M-RCNN, and let $P^{sem}$ denote the
dead tree class posterior probability image obtained from the U-Net.
Furthermore, let $\Omega \subset R^2$ be the image
plane, $I: \Omega \to R^d$ a vector-valued image, and $C$ an evolving contour
in the image $I$. The one-shape segmentation in the active contour model (ACM)
w.r.t.~shape and appearance priors $P(C),P(I|C)$ consists in finding a contour $C^{*}$ which
`optimally' partitions $I$ into disjoint interior and exterior regions such that
the probability $\mathcal{P}(C|I)\propto \mathcal{P}(I|C)\mathcal{P}(C)$ induced
by $C^{*}$ is minimized~\cite{CremersIJCV}:
\small
\begin{equation}
C^*=\argmin_C -\underbrace{\log \mathcal{P}(C|I)}_{\text{total energy}}=
\argmin_C [\,-\underbrace{\log \mathcal{P}(I|C)}_{\text{image term}}
- \underbrace{\log \mathcal{P}(C)}_{\text{shape term}}\,]
\end{equation}
\normalsize
Furthermore, the contour is parameterized by a vector of shape coefficients
$\bar{\alpha}$ and a offset vector $T=(t_x,t_y)$. A shape
generator $G(\bar{\alpha};T)$ is given, which instantiates the contour in
standard position and translates the center to $(t_x,t_y)$. The image term can
be interpreted as the pixel-wise cross entropy between the target class
posterior probability image $P^{sem}$ and the indicator function of the
contour's interior (see~\cite{isprs-archives-XLIII-B2-2020-717-2020} for
details). In our setting, we consider an arbitrary number of simultaneously
evolving contours $M$, each having its own shape coefficients and offset vector.
The image energy term is now defined as the cross entropy between the
set-theoretic union of all generated contours $G_i \equiv G(\bar{\alpha}_i,T_i),i
\in {1,\ldots,M}$ and the posterior probability image. Moreover, we introduce a new
term $E_{ovp}$ into the energy, which penalizes the total pairwise overlap
between evolving model shapes, to make sure they cover different regions of the
input image. We approximate the overlap between $G_i,G_j$ as the product
$\int_{\omega}G_i(\omega)G_j(\omega)$. The final energy formulation can be
written as:  
\small
\begin{equation}
\begin{split}
E(\bar{\alpha}_k,T_k,1 \leq k \leq M)&=-\gamma_{shp}\underbrace{\sum_{i=1}^{M}
\log \mathcal{P}(\bar{\alpha}_i)}_{\text{shape term}} -
\gamma_{img}\underbrace{\int_{\omega}U(G_{1,\ldots,M})[w]\log
P^{sem}(\omega)}_{\text{image term}}\\
&+\gamma_{ovp}\underbrace{\sum_{1 \leq i,j \leq
M}\int_{\omega}G_i(\omega)G_j(\omega)}_{\text{overlap term}}\\
& \,\,+E\,_{location}
\label{eq:energy-multi}
\end{split}
\end{equation}
\normalsize
In the above expression, the union operation $U(G_1,\ldots,G_M)$ can be implemented by
taking the pixel-wise maximum over all generated shapes $G_i$. However, since the
max function is not differentiable, we apply a smooth approximation
$U_\tau(x_1,\ldots,x_M) = \sum_i x_i e^{\tau x_i} / \sum_i e^{\tau x_i}$, where
$\tau$ is a positive constant. The coefficients $\gamma_{*}$ control the balance
of terms within the energy function. We utilize the eigenshape
model~\cite{Leventon2010,TSAI2019230} in the role of $G(\bar{\alpha};T)$, whereas the shape probability $P(\bar{\alpha})$ follows the kernel density
estimator model proposed by Cremers et al.~\cite{Cremers2007}. In practice, the
optimization requires good initial object positions and the object count
$M$. We utilize the centroids $p^{init}_{i}$ obtained from Mask RCNN in this
role. The evolving shape positions are constrained to lie within $\delta$ pixels
of $p^{init}_{i}$.

\vspace{-.2cm}
%
\section{Numerical Experiments}
\label{sec:exps}
\vspace{-.3cm}
%
%
%
%
\textbf{Data.} We use high resolution aerial images acquired by a flight campaign from the Bavarian Forest National Park in Germany with $10$ centimeter ground pixel resolution (see Appendix~\ref{sec:app-data-acq} for details).
We manually marked $201$ outlines of dead trees within the color infrared images
of a selected area in the National Park (Fig.~\ref{fig:refPolyManual}(a)) 
for training all components of the segmentation pipeline: the U-Net, the
Mask R-CNN and the active contour model. 
%
%
We employed a semi-automatic strategy for
acquiring dead tree crown polygon testing data. We applied the trained U-Net to a
new, previously unseen region of the National Park, and obtained the dead tree crown
per-pixel probability map. 
%
%
We subsequently manually partitioned a number of
connected components into individual tree crowns by applying split polylines to cut parts off the main polygon (Fig.~\ref{fig:refPolyManual}(c)),  
for a total of $750$ artificial tree crown polygons. 
These polygons were used to validate our approach and compare against the pure Mask R-CNN baseline.

\vspace{-.15cm}
\textbf{Training the models.}
%
%
For the U-Net, we followed the original architecture proposed
in~\cite{unet_paper}, and trained the network for $2000$ epochs on a total of $200$ patches of size $200 \times 200$ pixels.
%
%
%
We trained the Mask R-CNN on $70$ patches of size $256 \times 256$ until convergence of the validation loss curve ($100$ epochs).
(Implementation details can be found in Appendix~\ref{sec:app-implementation}).
%
%
%
The eigenshape model was learned from the training contours for the two CNNs, using $32$ top eigenmodes of variation and including rotated and flipped copies of the original polygons.
%
%

\begin{figure}[t]
\centering
\includegraphics[width=1\linewidth]{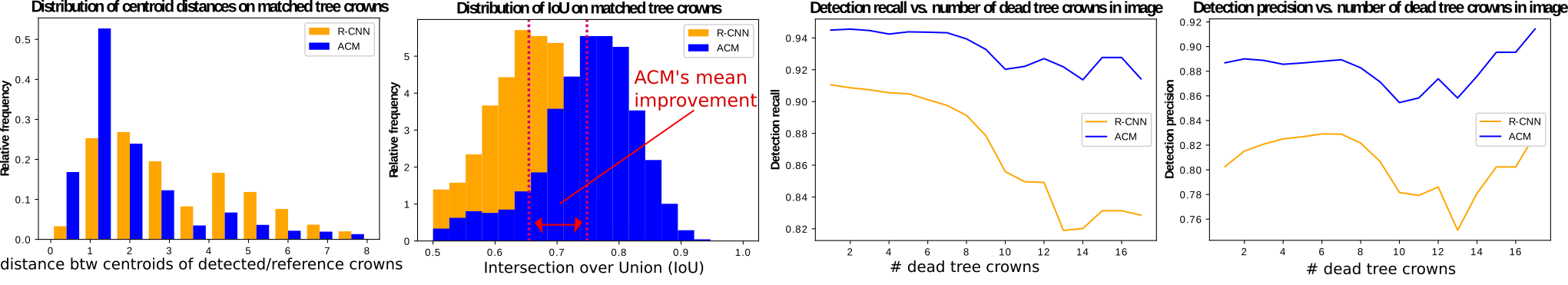}
\vspace{-.5cm}
\caption{Comparison of experimental results for baseline Mask R-CNN (\textcolor{orange}{orange}) and ACM (\textcolor{blue}{blue})
refinement methods. (a) distribution of centroid distances between matched
reference and detected dead tree crown polygons, (b) distribution of intersection over union on
matched crown polygons, (c) detection recall and (d) precision plotted
against number of dead trees in input image. The results show a significant
improvement on all metrics.}
\label{fig:results-graphs}
\end{figure}

\begin{figure}[t]
\centering
\includegraphics[width=.7\linewidth]{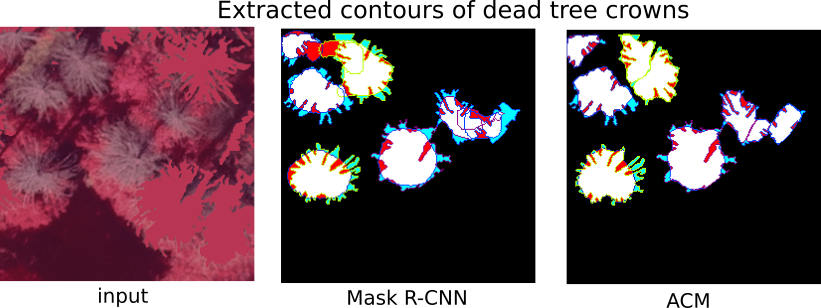}
\vspace{-.2cm}
\caption{
Comparison of ground-truth polygons with dead tree contours detected in the
input aerial image (left) by the baseline Mask R-CNN method (center) and the
proposed active contour refinement model (right).
White pixels represent areas of agreement, \textcolor{red}{red} pixels show false positive areas
(coverage by the model shape but not by ground-truth polygon), whereas \textcolor{cyan}{cyan}
pixels denote parts of ground truth polygons missed by the model. 
Our approach demonstrates a clear qualitative improvement in the quality of the refined contours, reflecting the quantitative results in Fig.~\ref{fig:results-graphs}.}
\label{fig:results-contours}
\end{figure}

\vspace{-.15cm}
\textbf{Contour retrieval performance.}
We ran several experiments comparing the quality of the extracted dead tree crown masks between the baseline method of Mask R-CNN and the
active contour model based refinement. 
To this end, the aforementioned $750$ dead
tree crowns were distributed into $285$ images of dimensions $256 \times 256$ (same as
training patches). 
We executed the pipeline from~Fig.\ref{fig:pipeline} on the test images 
until convergence, 
yielding refined contours. 
%
%
%
To solve the (box-)constrained continuous energy minimization problem from Eq.~\eqref{eq:energy-multi}, we used the L-BFGS method.
To assess the quality of both sets of masks, we used the following metrics: (i) mean centroid distance between reference and detected tree crown masks
mean, (ii) Intersection over Union (IoU)~\cite{jaccard1901etude} of detected vs.~reference polygons, and
%
%
(iii) precision and recall at IoU $\le 0.5$.
%
%
The results are visualized in ~Fig.\ref{fig:results-graphs}. The true centroids
of the dead tree crowns can be approximated by the centroids found by Mask R-CNN
very well (average deviation of $3.4$ pixels) and thus serve as good seed
points for the ACM contour model. The ACM refinement further improved this
value by ca.~$1$ pixel to $2.4$ pixels, as shown
in~Fig.\ref{fig:results-graphs}. On average, the IoU improved by ca.~$9$
percentage points (pp) after refinement, leading to an increase in precision and
recall by, respectively $8$ and $3.5$ pp. Moreover, we observe that as the
number of dead trees present within the image increases, the detection recall for the ACM
refinement does not drop as quickly as for the baseline method. There were a
total of $2041$ vs. $1917$ dead trees reported respectively by Mask R-CNN and ACM
methods. Sample dead tree crown contours from our ACM approach and Mask
R-CNN are shown in~Fig.\ref{fig:results-contours}.


%

\vspace{-.2cm}
\section{Conclusions and Discussion}
\label{sec:disc}
\vspace{-.3cm}
%
%
Dead trees are a key indicator of overall forest health, biodiversity and are a crucial component of forest carbon dynamics that are heavily influenced by climate change, insects, and fungi. 
In order to aid in understanding e.g. dead wood decomposition, this work proposed a hybrid of two convolutional neural networks (U-Net and Mask R-CNN) with a novel active contour model (ACM) for precise dead tree contour segmentation.

Our numerical experiments comparing our (ACM) approach to Mask R-CNN as a baseline
show that although the latter yields good estimates of the number and
location of dead trees in an image, the alignment of the detected
contour with the true dead crown is poor. On the other hand, applying the ACM based contour refinement can significantly
improve this alignment (by $9$ pp on average), as measured by the overlap (IoU). 
%
%
%
Furthermore, these experiments show that ACM is more robust in the
presence of more difficult scenarios as measured by the number of dead trees
present in the image. 

Future work includes incoorporating contour shape priors that can capture even more fine details of the tree crowns (e.g. generated by GANs) than can the current eigenshape prior.
Another focus will be to use the proposed method to improve estimates of dead wood decay rates by means of, e.g.~temporal change detection, with which to ideally form models of decay dynamics dependent on different factors, e.g. geographical location and tree species.
These additions will further help monitor and understand forest ecosystem health and biodiversity, and the role of dead wood and its impacts on and from our rapidly changing climate.

\small
\bibliographystyle{ieeetr}
\bibliography{ci_references}

\newpage
\appendix

\section{Appendix: Experiments}
\label{appendix}
\subsection{Implementation of U-Net and Mask R-CNN}
\label{sec:app-implementation}
The tensorflow implementation of the U-Net in~\cite{akeret2017radio}
was adapted to support masking out irrelevant parts of the images
in the training phase. We used the original architecture proposed
by~\cite{unet_paper}, and trained the network for $2000$ epochs on a total of $200$
patches of size $200 \times 200$ pixels.

We used the implementation of Mask R-CNN in~\cite{matterport_maskrcnn_2017}
publicly available on Github. Image augmentation was applied in the form of
horizontal and vertical flipping as well as rotation by $90$, $180$ and $270$ degrees.
The optimization on $70$ patches of size $256 \times 256$ was conducted until convergence
of the validation loss curve ($100$ epochs). 
%
%
Before training the Mask R-CNN, all images were inspected for dead tree crowns which were not labeled. Such tree crowns were overwritten with a neutral color within the image so that detection metrics may be reliably computed (all dead trees detectable within the image are annotated with ground-truth labels).

The eigenshape model was learned from the training contours for
the two CNNs, using $32$ top eigenmodes of variation 
 and including rotated 
and flipped copies of the original polygons.
The tree crown masks were aligned according to their centroid wihin a 92x92 pixel frame, corresponding to the largest object we wish to detect (crown diameter of 9.2).

\subsection{Further refined contour examples}
\label{sec:app-examples}
\begin{figure*}[h!]
  \subfigure[]{\includegraphics[width=.31
  \columnwidth]{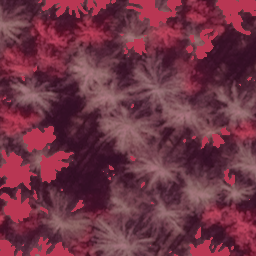}}\hfill
  \subfigure[]{\includegraphics[width=.31
  \columnwidth]{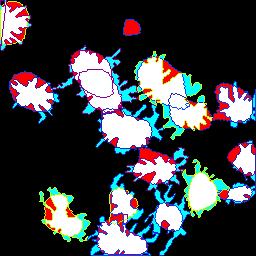}}\hfill
  \subfigure[]{\includegraphics[width=.31
  \columnwidth]{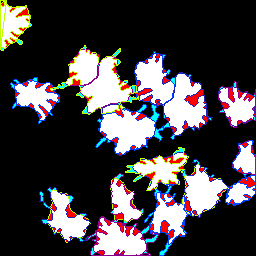}}\hfill
  \caption{Comparison of ground-truth polygons with dead tree contours detected in the
input aerial image (left) by the baseline Mask R-CNN method (center) and the
proposed active contour refinement model (right).
White pixels represent areas of agreement, red pixels show false positive areas
(coverage by the model shape but not by ground-truth polygon), whereas cyan
pixels denote parts of ground truth polygons missed by the model. We see an
improvement in the quality of the refined contours.}
  \label{fig:sampleCrowns2}
\end{figure*}

\section{Appendix: Data}
\label{sec:app-data}
\subsection{Data acquisition}
\label{sec:app-data-acq}
Color infrared images of the Bavarian Forest National Park, situated in
South-Eastern Germany ($49^\circ 3' 19''$
N, $13^\circ 12' 9''$ E), were acquired in the leaf-on state during a flight campaign carried out in June 2017
using a DMC III high resolution digital aerial camera.The mean above-ground flight height was ca. 2300 m, resulting in a pixel resolution of 10 cm on the ground. The images
contain 3 spectral bands: near infrared, red and green. 

\subsection{Testing and training data}
\label{sec:app-data-traintest}
\begin{figure}[h!]
  \begin{center}
  \subfigure[]{\includegraphics[width=.5
  \columnwidth]{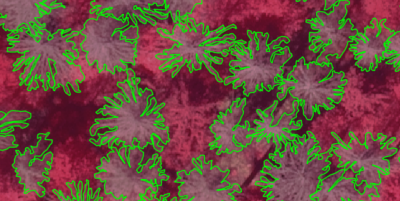}}\hfill
  \subfigure[]{\includegraphics[width=.5
  \columnwidth]{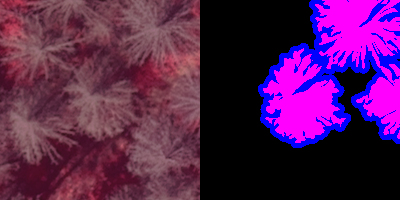}}\hfill\\
    \begin{center}
  \subfigure[]{\includegraphics[width=.5
  \columnwidth]{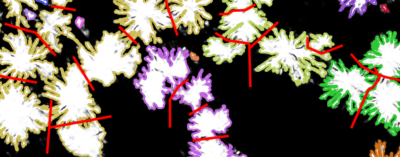}}\hfill
    \end{center}
  \caption{(a) manually marked dead tree crown polygons within CIR image, (b)
  left: input CIR image patch for U-Net training, right: matching label mask:
  positive and negative class labels indicated by magenta and blue,
  respectively, black mask regions do not count toward training loss, (c)
  manually drawn split polylines (in red) to separate connected components into
  individual tree crowns.}
  \end{center}
  \label{fig:refPolyManual}
\end{figure}

\begin{figure}[h!]
\centering
\includegraphics[width=0.5\linewidth]{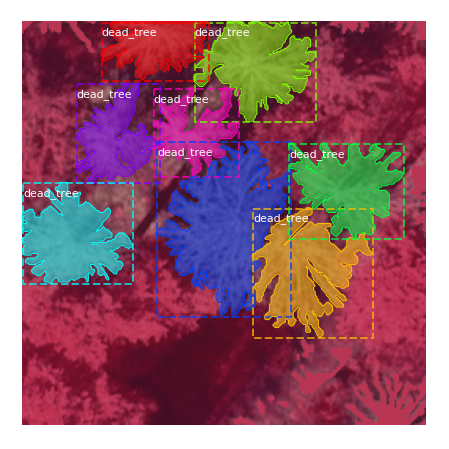}
\caption{Reference polygons marked on a 256x256 image for Mask R-CNN training.}
\label{fig:rcnnTraining}
\end{figure}

As mentioned in the Experiments section~\ref{sec:methods}, 
high resolution aerial data were acquired by a flight campaign from the Bavarian Forest National Park with $10$ centimeter ground pixel resolution (see Appendix~\ref{sec:app-data-acq} for details).
We manually marked 201 outlines of dead trees within the color infrared images
of a selected area in the National Park (see Fig.~\ref{fig:refPolyManual}a).
These manually marked polygons were utilized for the purpose of training all
components of the segmentation pipeline: the U-Net, the Mask R-CNN and the active contour model. For training the U-Net, we prepared
patches of size 200x200 containing the input color infrared image and a pixel
mask representing the labeled polygon regions. Also, we constrained the
negative class labels to at most 5 pixels away from labeled dead tree polygons,
to account for the fact that not all dead tree crowns in the processed images
were labeled (Fig.~\ref{fig:refPolyManual}b). For training the Mask R-CNN, we
utilized 70 patches of size 256x256 with marked individual instances as
input (see Fig.~\ref{fig:rcnnTraining}). Finally, the binary masks of individual
marked tree crown polygons were used as a basis for learning the active contour
model.

We employed a different, semi automatic strategy for
acquiring dead tree crown polygon testing data. We applied the trained U-Net to a
new, previously unseen region of the National Park, and obtained the dead tree crown
per-pixel probability map. Connected component segmentation was then applied on
pixels of the image classified as dead trees. As the test area contained many
overlapping and adjacent dead trees, the connected components obtained from
this step usually did not represent only single trees, but rather collections
of several dead tree crowns. We subsequently manually partitioned a number of
connected components into individual tree crowns by applying split polylines to
successively cut parts off the main polygon (Fig.~\ref{fig:refPolyManual}c). We
found this approach to be less time consuming than manually drawing the entire
polygons. We obtained a total of 750 artificial tree crown polygons this way.
They were utilized for validating our approach and for comparison against the
pure Mask R-CNN baseline.

\end{document}